\documentclass[conference]{IEEEtran}
\IEEEoverridecommandlockouts
\usepackage{cite}
\usepackage{amsmath,amssymb,amsfonts}
\usepackage{algorithmic}
\usepackage{graphicx}
\usepackage{textcomp}
\usepackage{xcolor}

\usepackage[utf8]{inputenc} % allow utf-8 input
\usepackage[T1]{fontenc}    % use 8-bit T1 fonts
\usepackage{url}            % simple URL typesetting
\usepackage{booktabs}       % professional-quality tables
\usepackage{amsfonts}       % blackboard math symbols
\usepackage{nicefrac}       % compact symbols for 1/2, etc.
\usepackage{microtype}      % microtypography
\usepackage{xcolor}         % colors
\usepackage{graphicx}
\usepackage{subfigure}
\usepackage{pifont}
\usepackage{multirow}
\usepackage{url}
\usepackage{makecell}
\usepackage{amsmath}

\usepackage{amssymb}

\usepackage{algorithm}
\usepackage{algorithmic}

\usepackage{hyperref}

\def\BibTeX{{\rm B\kern-.05em{\sc i\kern-.025em b}\kern-.08em
    T\kern-.1667em\lower.7ex\hbox{E}\kern-.125emX}}
\begin{document}

\title{LEFL: Low Entropy Client Sampling in Federated Learning \\
%{\footnotesize \textsuperscript{*}Note: Sub-titles are not captured in Xplore and should not be used}
%\thanks{Identify applicable funding agency here. If none, delete this.}
}

\author{\IEEEauthorblockN{Waqwoya Abebe}
\IEEEauthorblockA{\textit{Computer Science} \\
\textit{Iowa State University}\\
Ames, IA \\
wmabebe@iastate.edu}
\and
\IEEEauthorblockN{Pablo Munoz}
\IEEEauthorblockA{\textit{Intel Labs} \\
\textit{Intel Corporation}\\
Santa Clara, CA \\
pablo.munoz@intel.com}
\and
\IEEEauthorblockN{Ali Jannesari}
\IEEEauthorblockA{\textit{Computer Science} \\
\textit{Iowa State University}\\
Ames, IA \\
jannesar@iastate.edu}
% \and
% \IEEEauthorblockN{4\textsuperscript{th} Given Name Surname}
% \IEEEauthorblockA{\textit{dept. name of organization (of Aff.)} \\
% \textit{name of organization (of Aff.)}\\
% City, Country \\
% email address or ORCID}
% \and
% \IEEEauthorblockN{5\textsuperscript{th} Given Name Surname}
% \IEEEauthorblockA{\textit{dept. name of organization (of Aff.)} \\
% \textit{name of organization (of Aff.)}\\
% City, Country \\
% email address or ORCID}
% \and
% \IEEEauthorblockN{6\textsuperscript{th} Given Name Surname}
% \IEEEauthorblockA{\textit{dept. name of organization (of Aff.)} \\
% \textit{name of organization (of Aff.)}\\
% City, Country \\
% email address or ORCID}
}

\maketitle

\begin{abstract}
  Federated learning (FL) is a machine learning paradigm where multiple clients collaborate to optimize a single global model using their private data. The global model is maintained by a central server that orchestrates the FL training process through a series of training rounds. In each round, the server samples clients from a client pool before sending them its latest global model parameters for further optimization. Naive sampling strategies implement random client sampling and fail to factor client data distributions for privacy reasons.  Hence we propose LEFL \footnote{Github: \href{https://github.com/wmabebe/lefl}{ \textbf{L}ow \textbf{E}ntropy Client Sampling in \textbf{F}ederated \textbf{L}earning}}, an alternative sampling strategy by performing a one-time clustering of clients based on their model's learned high-level features while respecting data privacy. This enables the server to perform stratified client sampling across clusters in every round. We show datasets of sampled clients selected with this approach yield a low relative entropy with respect to the global data distribution. Consequently, the FL training becomes less noisy and significantly improves the convergence of the global model by as much as 7.4\% in some experiments. Furthermore, it also significantly reduces the communication rounds required to achieve a target accuracy.
\end{abstract}


\section{Introduction}

Federated learning~(FL) \cite{mcmahan2017fedavg, karimireddy2020scaffold} is a machine learning technique used to train machine learning models on decentralized client data. The goal of FL is to train a single global model on collective client data while maintaining client data privacy. FL is organized and orchestrated by a central server that is responsible for managing the global model. The global model is trained in a series of rounds whereby in each round, a sample of clients download the global model parameters, optimize them locally with their private data and upload them back to the server. The server, on its part, aggregates the uploaded model parameters to update the global model for the next round. This iterative process helps maintain client privacy since only parameters/gradients are communicated with the server. 

Despite the overall simplicity of the training process, FL poses several challenges such as data and system heterogeneity \cite{zawad2021curse, li2020federated}, which can degrade the global model's performance. Additionally, communication overhead \cite{luping2019cmfl} also poses a major challenge considering the bandwidth limitations of participating client edge devices. Data heterogeneity refers to the imbalance of features, label distribution and/or sample quantities across client datasets \cite{li2022federated}. Such imbalance causes gradient drift \cite{yu2021spatl} between local updates and server updates (which occurs after aggregation). Gradient drift can arise from client model's overfitting local biased data, bias in aggregation \cite{cho2022towards} and covariate shift \cite{gao2019privacy}. In turn, gradient drift introduces noise in the global model's training process and negatively impacts its overall performance.

%Several algorithms and methods have been proposed to mitigate these issues and are discussed in more detail in the Related Works section. 
%For instance, client clustering techniques have been proposed to address these challenges by grouping clients with similar data distributions, network characteristics, or hardware capabilities together to boost client cooperation and improve the training process.

% \todo[inline]{The first few sentences of the following paragraph might be good to move them to Related Works. Mention limitations and then "In this paper...}

%Several model aggregation techniques have been proposed in federated learning, each with strengths and limitations.

In every round, the FL server samples a subset of clients that will optimize the global model. Uniform sampling \cite{li2019convergence} is a common client sampling technique, but it fails to account for client heterogeneity. In particular, due to client data privacy, the server is unable to sample clients in a representative manner where the collective dataset of the sampled clients resembles the global data distribution. As a result, sampled clients often contain biased data and inevitably upload biased gradients exacerbating gradient drift. Although some works introduce stratified sampling, importance sampling \cite{rizk2021optimal} and adaptive sampling \cite{luo2022tackling}, they mostly focus on system heterogeneity or client data size variations. Hence, they fail to address variability in client data distributions (data heterogeneity). This begs the question, can we implement representative client sampling to address data heterogeneity without infringing on client data privacy?

In this paper, we propose a pre-processing step to cluster FL clients based on the similarity of their models latent layers. We sandwich this clustering step between the first and second training rounds. Initially, all FL clients download a small unlabeled public dataset from a different distribution. After conducting a single FL training round on their private dataset, clients use their models to predict the soft-labels of the public dataset. All soft-labels are then uploaded to the server to construct a similarity matrix \cite{abebe2023optimizing} which is  used to classify the clients in an unsupervised manner. We demonstrate how this clustering approach groups together models that have learned similar high-level features. Clustering enables the server to conduct stratified client sampling across the clusters such that the sampled clients data yields low relative entropy with respect to the global data distribution. Consequently, the FL training process becomes less noisy, thus improving the global model's performance. In summary:
\begin{itemize}
    \item We propose a client clustering approach based on the models high-level learned features.
    \item We provide empirical evidence to justify how this scheme allows sampling clients with low relative entropy with respect to the global data distribution.
    \item We demonstrate the effectiveness of low entropy client sampling in various FL experiments in terms of global model accuracy, communication efficiency, and convergence speed. 
\end{itemize}

The remainder of the paper is structured as follows: We present federated learning and aggregation techniques in the Related Works section. The Methodology section describes our proposed client clustering technique, including the clustering algorithm. The Evaluation section presents the experimental setup and empirical evidence to justify the proposed approach. Moreover, we evaluate the proposed technique with various datasets and models against strong baselines. Finally, we summarize our main contributions in the Conclusion.

\section{Related Works}\label{sec:related_works}

In recent years, several schemes have been proposed to address data heterogeneity in FL. These include, modifying the aggregation, clustering clients, varying model architectures and utilizing knowledge distillation or other techniques. Below, we discuss some aggregation techniques and sampling schemes.

\subsection{Aggregation Techniques}

One way to tackle data heterogeneity in FL, is to tweak the aggregation method. The first FL work \cite{mcmahan2017fedavg} proposed FedAvg, a direct averaging of client parameters. Later, FedProx \cite{li2020federated}, proposed adding a proximal term, $\frac{\mu}{2} || x_i - x^{(t,0)} ||^2$, to the client's local cost function. This extra term helps narrow the gap between the local and global gradients. SCAFFOLD \cite{karimireddy2020scaffold}, proposes maintaining control variates on both the server and clients which can be used to control local gradient drift. FedNova \cite{wang2020tackling}, proposes a global learning rate adaptation coefficient based on the gradient variance of the aggregated model. Each client then adjusts its learning rate according to the server updates.

Several knowledge distillation-based works have also been proposed to tackle the effect of data heterogeneity. To address the issue of communication overhead, one approach has focused on applying model compression techniques \cite{shah2021model, sattler2019robust, xu2020ternary}. For instance in \cite{yu2022spatl}, authors proposed reducing the communication overhead by only communicating the encoder component of the model. Furthermore, they implement a local reinforcement learning (RL) agent to filter salient parameters of the encoder for communication. Similarly, knowledge distillation methods have been proposed to help reduce communication overhead \cite{li19fedmd,he20fedgkt,seo20fedkd,yu2022resource}. These methods usually rely on communicating a smaller knowledge network that is optimized by clients locally.

Our proposed client sampling strategy is implemented over the basic FedAvg approach. However, it is quite simple to adapt it for any of the above aggregation schemes since the only additional step we are proposing is a pre-clustering of clients before full participation plus stratified sampling in each round.

\subsection{Sampling Techniques}

The initial federated learning work (FedAvg) \cite{mcmahan2017fedavg}, proposed sampling clients uniformly in each round. Despite the simplicity of this approach, it was later shown in SCAFFOLD \cite{karimireddy2020scaffold} that random sampling is biased and can cause client gradient drift. To tackle this issue, FedProx \cite{li2020federated} proposed sampling clients from the multinomial distribution where clients are sampled proportional to the size of their dataset. These approach however, only factored in the dataset size of the participating clients. 

In \cite{li2019convergence}, authors propose a sampling the first $m$ client updates. This prioritizes powerful clients or clients with fewer samples. In \cite{fraboni2021clustered}, authors provide two algorithms for clustering clients before applying sampling. The first algorithm clusters clients based on local dataset size, while the latter clusters clients based on gradient similarity. They showed that such extra client cluster information is able to improve the sampling techniques. However, the extra gradient information is as large as the uploaded parameters. \cite{duan2019astraea} introduces data augmentation and an extra component called a mediator with knowledge of local class label distribution. \cite{cho2020client} and \cite{goetz2019active} propose selecting  clients that exhibit larger loss. However, this approach does not guarantee optimal performance.

\cite{yang2021federated} implemented an estimation scheme to determine the label distribution among the participating clients. Afterwards, they implement a greedy algorithm to pick the set of clients on each round. However, this work expects the server to posses a balanced dataset which may not always be practical.  More recently, \cite{zhang2023fed} proposed a heterogeneity-aware client sampling mechanism to address class imbalance. They propose a measure, ``Quadratic Class-Imbalance Degree'' (QCID), to determine the heterogeneity of sampled clients and implement a greedy algorithm to select client samples that minimize this value. Since QCID is computed based on the clients underlying data distribution. In this approach, clients will have to transmit their data distribution information via third party utilities such as fully homomorphic encryption (FHE), requiring a computationally costly service that might not be readily available to participating clients.

While the naive uniform sampling strategy potentially suffers from gradient drift, the improved unbiased sampling strategies mostly only factor the proportion of samples owned by clients when factoring their sample ratio. Works like \cite{yang2021federated}, \cite{zhang2023fed} that factor client data distribution require extra assumptions or expensive procedures to compute distribution similarities among clients. In our work, we propose clustering clients based on their models high-level learned features before applying stratified random sampling over the clusters. Moreover, once the clustering is complete, the sampling strategy can easily be augmented with unbiased sampling techniques proposed in prior work. Our experiments show that our proposed scheme significantly improves the performance of the global model compared to the state-of-the-art FL baselines.

\section{Methodology}
\label{sec:method}

\begin{figure*}
\centering
  \includegraphics[width=0.7\linewidth]{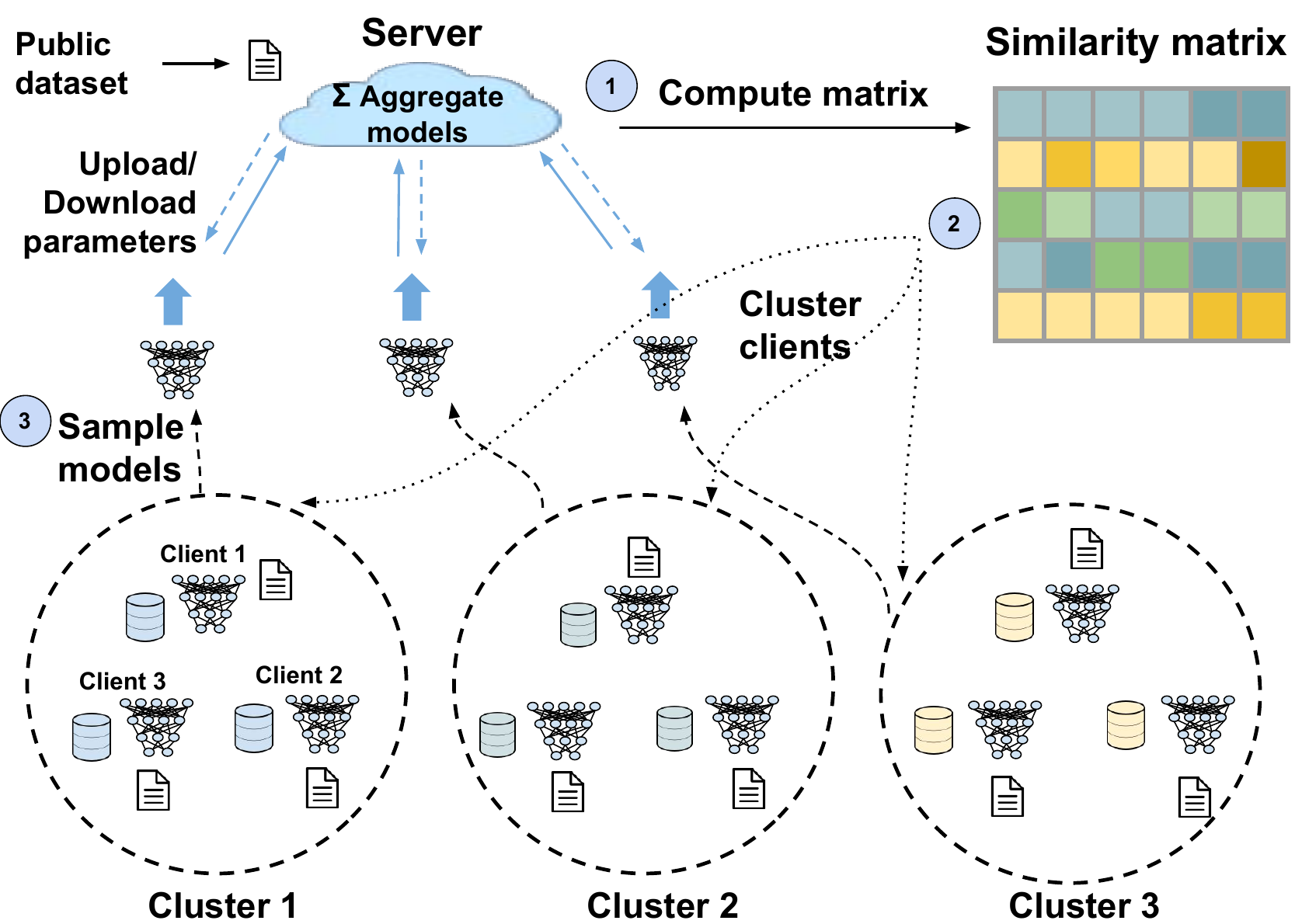}
\caption{Server side computation after receiving soft-labels from clients (Algorithm \ref{alg:preprocess}). The server will construct a similarity matrix and use it to cluster the clients. Afterwards, it will conduct stratified client sampling across clusters in each FL round. }
\label{fig:overview}%
\end{figure*}

The objective of FL is to optimize a single global model using decentralized client data. In each FL round, the global model's parameters are sent to a selected subset of clients for local optimization. The sampled clients optimize their copy of the global model's parameters using their local data for a predefined number of epochs before uploading them back to the server.  The server, in turn, aggregates the uploaded parameters to update the next version of the global model as shown in (\ref{eq:fedavg}).

The baseline FedAvg algorithm  proposes updating the global model parameters $w$, by aggregating the latest local parameters, $w^t_i$,  uploaded by a client $C^i$ in the current round $t$:

\begin{equation}
    w^{t+1} = \sum_{i\in s} \frac{|D_i|}{|D|} w^{t}_i
    \label{eq:fedavg}
\end{equation}

where $|D_i|$ is the size of the local dataset $D_i$, $|D|$ is the size of the global dataset $D$, $s$ is the set of sampled clients and $t$ is the FL communication round.

In practice, FL clients are expected to possess non-iid (non-independent and identically distributed) data. Non-iid data can manifest as feature skew, label skew or both. In classification problems for instance, the label distributions of clients is expected to vary. Intuitively, such skews introduce noise into the training process, as variation in client data distribution causes client gradient drift. This can lead to slower convergence and degrade performance of the global model. In contrast, in the iid case where data is evenly distributed across all nodes, performing uniform client sampling will suffice for sampling clients that possess representative (iid) data.  Here, we define a dataset $D_s$ to be representative if it has a similar distribution to the global dataset $D$, i.e., the distribution of the union of datasets belonging to the sampled clients is similar to the global data distribution. This helps the model converge faster and achieve better performance.

 In the label distribution skew case for instance, the similarity between datasets can be measured as the distance between the probability distribution of $D$ and $D_s$, e.g., using the Kullback–Leibler divergence \cite{KLDivergence} (\ref{eq:kldiv}) or other statistical distance metric. Under such circumstance, selecting a subset of clients which contain a balanced number of labels in each round will help reduce aggregation noise. This is tantamount to selecting clients whose collective data is iid. In the non-iid case however, random sampling of clients is unlikely to yield such a representative subset of client data. Since the server is unaware of client data, it has no way of directly measuring the representativity of the datasets belonging to the clients it sampled. 

With this insight, we want to implement a representative client sampling strategy that can select a subset of clients in each federated round where the selected clients posses representative underlying data. To circumvent server blindness, we propose using the model's learned high-level features as an indication of underlying client dataset. \textbf{In particular, we show that selecting clients that have learned a diverse set of high-level features is a good substitute for selecting clients based on their underlying data distribution itself.} In this section, we propose an affordable scheme for the server to indirectly infer the models learned high-level features without directly accessing clients data or requiring (all) clients to upload their models. While the former approach violates the clients data privacy, the latter becomes infeasible as the system scales. Afterwards, the server uses this information to cluster similar clients together and apply a stratified sampling across the clusters yielding sampled clients whose combined dataset has a low relative entropy with respect to the global data distribution.

\begin{algorithm}[tb]
\caption{Extra pre-processing steps}
\label{alg:preprocess}
\begin{algorithmic}

\STATE \textbf{Client side}
    
    \STATE \quad 1. Download unlabeled public dataset (initial contact)
    \STATE \quad 2. Train global model for $E_p$ epochs on private data
    \STATE \quad 3. Predict soft-labels of public data
    \STATE \quad 4. Upload soft-labels

\STATE \textbf{Server side}

    \STATE \quad 1. Construct similarity matrix
    \STATE \quad 2. Cluster clients
    \STATE $\circlearrowright$ Perform stratified client sampling  for all rounds

\end{algorithmic}
\end{algorithm}

% \begin{algorithm}[tb]
% \caption{Example algorithm}
% \label{alg:algorithm}
% \textbf{Input}: Your algorithm's input\\
% \textbf{Parameter}: Optional list of parameters\\
% \textbf{Output}: Your algorithm's output
% \begin{algorithmic}[1] %[1] enables line numbers
% \STATE Let $t=0$.
% \WHILE{condition}
% \STATE Do some action.
% \IF {conditional}
% \STATE Perform task A.
% \ELSE
% \STATE Perform task B.
% \ENDIF
% \ENDWHILE
% \STATE \textbf{return} solution
% \end{algorithmic}
% \end{algorithm}

% In \cite{zhang18dml}, authors proposed a mechanism of deep mutual learning between a model pair by minimizing the KL-divergence loss between the model's soft-label outputs in addition to the models' cross-entropy loss. This process was used to narrow the gap between learned representation of the two models and transfer/condense knowledge between the networks. In \cite{abebe2022addressing}, authors compute a similarity matrix in order to construct a locally heterogeneous decentralized learning topology. Our method, which is inspired by these works, seeks to utilize the KL-divergence as a means to compare the similarities between learned high-level feature representations. Afterwards, we use this information to cluster the clients in order to perform a stratified client sampling in FL. 

Prior to initializing the FL training, the server prepares a public dataset $D_p$ with a limited number of samples. This dataset is shipped to the clients upon initial download of the global model. After receiving the global model and public data, clients train their copies of the global model for a limited number of epochs $E_p$ on their private data and use their trained model to predict the labels of the public dataset. Afterwards, clients will upload the soft-labels outputted on the public dataset back to the server. Having received the soft-labels, the server computes pairwise KL-divergence values between all pairs of soft-labels to create a similarity matrix as shown in Figure~\ref{fig:overview}. 

Given a set of clients $C$ where $|C| = n$, the similarity matrix contains KL-divergence values $\alpha^i_j \quad \forall \: C^i,C^j \in C$. If two clients $C^{i}, C^{j}$ have similar rows $\alpha^i_k, \alpha^i_k \quad \forall  \: k \in \{1..n\} \setminus \{i,j\}$, we observe that they are similarly divergent from other models in the matrix. This meta information can be used to cluster the clients using an unsupervised clustering algorithm, e.g., KMeans \cite{Hartigan1979}. As such, the server groups the clients into $log(n)$ non-overlapping clusters as shown in Figure~\ref{fig:overview}. Once the clients are clustered, the server can now commence the FL process by applying a stratified sampling over the clusters.

To understand why our similarity matrix can help cluster similar clients together, suppose there are N clients that are training a global deep learning classification model (e.g., ResNets) in a federated manner. Each client $C^i$ trains its copy of the global model $m_i$ locally on its private data $D_i$ for a limited number of epochs (e.g., 10). Afterwards, it uses $m_i$ to predict labels on an unlabeled public test dataset $D_p$. Let $p_i$ and $p_j$ be the soft-label probability distributions output by models $m_i$ and $m_j$ on the public dataset $D_p$. The KL-divergence between $p_i$ and $p_j$ can be written as:

\begin{equation}
    \alpha^i_j = KL(p_i || p_j) = \sum_{c=1}^K p_{i,c} \log\left(\frac{p_{i,c}}{p_{j,c}}\right)
    \label{eq:kldiv}
\end{equation}

where $K$ is the number of classes in the classification problem. We make two conjectures why this approach helps group similar clients together. %Let $m_i$ and $m_j$ be two identical models trained on two datasets $D_i$ and $D_j$ with different distributions. Assume the same model initialization, hyper-parameter settings and optimization algorithms are used to train $m_i$ and $m_j$, and the models are trained for the same number of epochs. Let $p_i$ and $p_j$ be the soft-labels produced by $m_i$ and $m_j$ respectively, when fed an unlabeled test data. 

\paragraph{Assumption 1.} All client models are trained on the same task with the same model initialization, hyper-parameter settings and optimization algorithms, and the models are trained for the same number of epochs.

\paragraph{Assumption 2.} The global dataset $D$ (collective client dataset) is composed of a limited number of labels (e.g., 10). Each client's local dataset contains an arbitrary number of unique labels (labels $\in$ $D_i$ $\subseteq$ labels $\in$ $D$ ) and an arbitrary number of samples for each label.

High-level features can be represented as a function of the input data and the model's parameters. Let's denote the input data as $x$, and the parameters of the model as $\theta$. The high-level features can be denoted as $H(x; \theta)$, which indicates that the features are obtained by applying a transformation $H$ to the input $x$ using the model's parameters $\theta$.

\paragraph{Conjecture 1:} If the KL-divergence, $\alpha^i_j$,  between $p_i$ and $p_j$ is small, then we can expect that the high-level learned representation of models $m_i, m_j$ could be similar.

\begin{equation}
    \alpha^i_j + \alpha^j_i < \epsilon_1   \rightarrow \left\|H_i(x; \theta_i) - H_j(x; \theta_j)\right\| < \epsilon_0
    \label{eq:conj_1}
\end{equation}

where $\epsilon_0$ and $\epsilon_1$ are small threshold values.

\paragraph{Conjecture 2:} If two rows in the KL-based similarity matrix, $\alpha^i_k, \alpha^j_k$, have similar values, then models $m_i$ and $m_j$ have learned similar high-level feature representations via relativity with the same subset of models.

\begin{equation}
     \frac{1}{N} \sum_{k} |\alpha^i_k - \alpha^j_k| < \epsilon_2 \rightarrow \left\|H_i(x; \theta_i) - H_j(x; \theta_j)\right\| < \epsilon_0
    \label{eq:conj_2}
\end{equation}

where $\epsilon_2$ is a threshold value.

This is because the soft-label distribution is determined by the learned feature representations and the decision boundaries in the classification task. If two identical models trained on datasets with differing data distributions produce similar soft-label distributions, then it suggests that they have learned similar decision boundaries, which in turn indicates that they have learned similar feature representations. In the evaluation section we first empirically test the conjectures and compare the proposed scheme against baseline FL aggregation methods.

\section{Evaluation}\label{sec:evaluation}

In this section we present experiments to demonstrate the viability of the proposed scheme. We first conduct empirical tests to evaluate our conjectures. Afterwards, we run FL experiments to compare our clustering scheme against strong baselines.

\subsection{Testing Conjectures}

\begin{figure}

    \centering
    \includegraphics[width=1\linewidth]{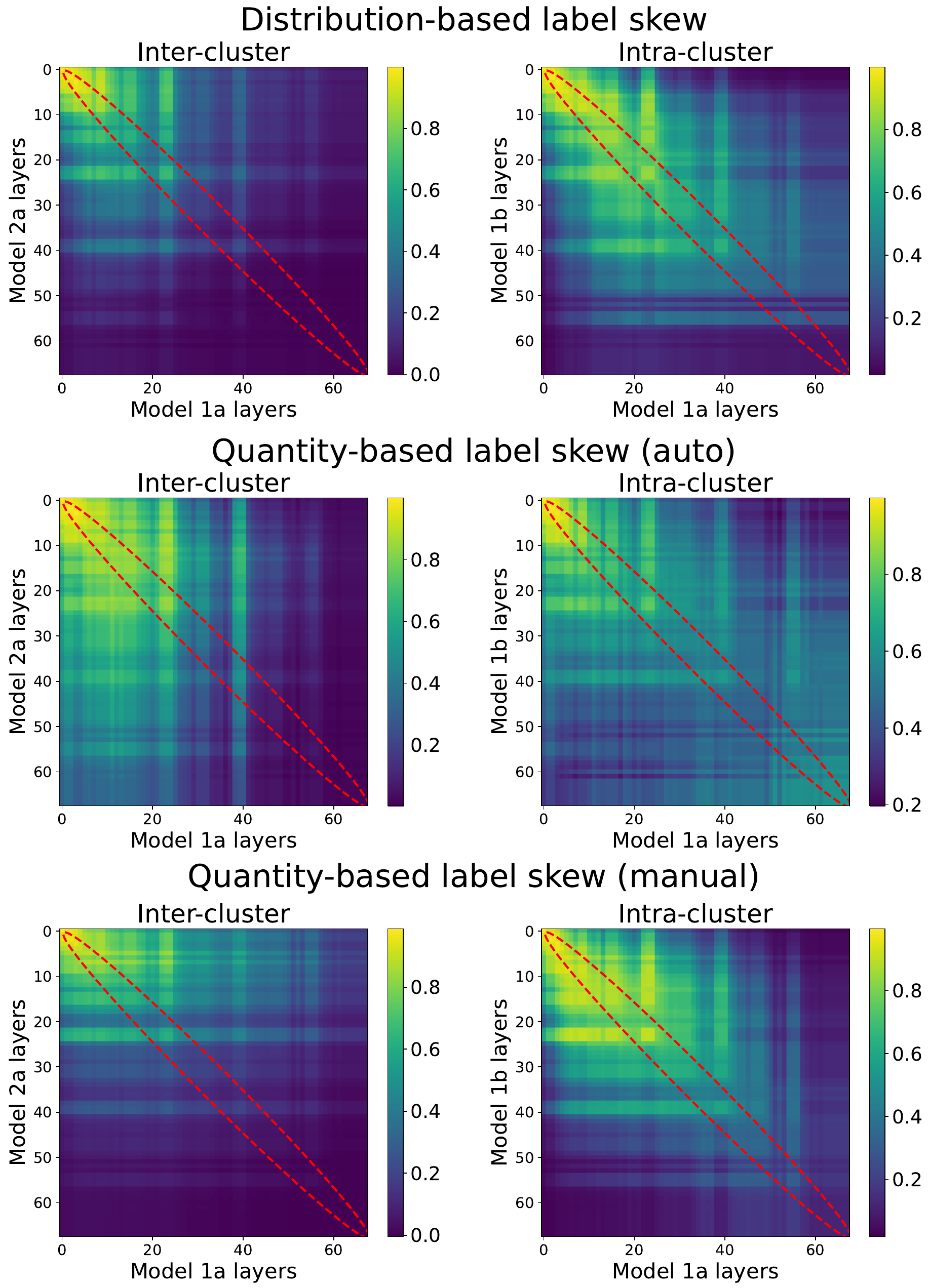}
    
    \caption{Comparing CKA values for every layer of a pair of randomly selected models. Models 1a and 1b belong to the same cluster, whereas model 2a comes from a different cluster. Red diagonal highlights region where model layers intersect.}
    \label{fig:cka}
\end{figure}

\begin{figure}[htb]
    \centering
    \includegraphics[width=.8\linewidth]{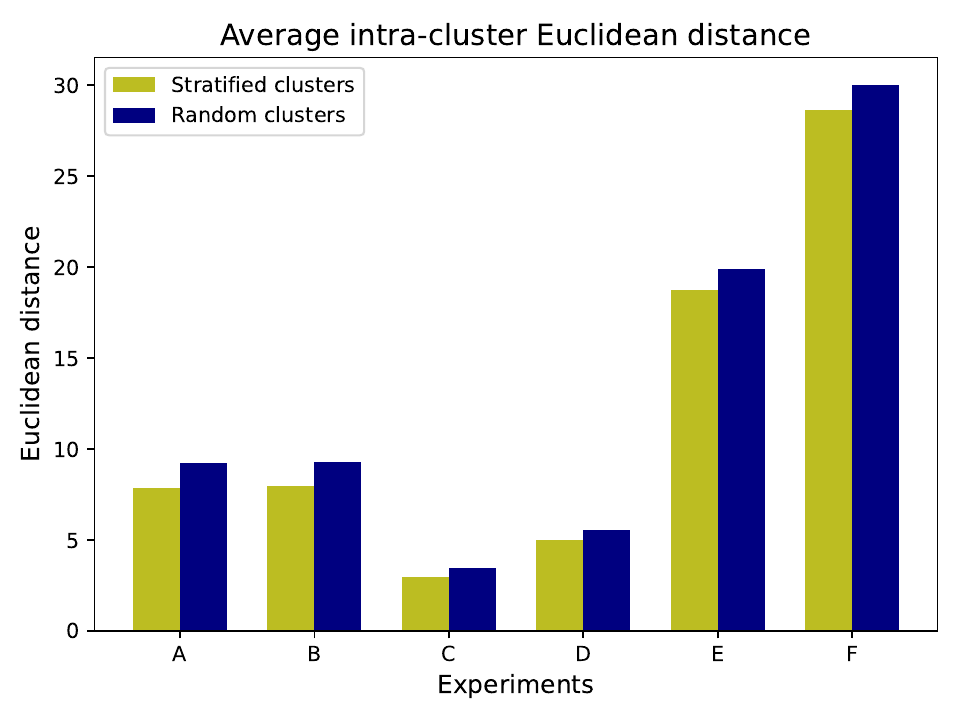}
    \caption{Average Euclidean distance of latent space vectors among clients within stratified clusters vs. random clusters. Details of experiments A - F will be discussed in supplementary materials.}
    \label{fig:euclidean}
\end{figure}

\begin{figure*}[htb]
    \centering
    \includegraphics[width=0.75\linewidth]{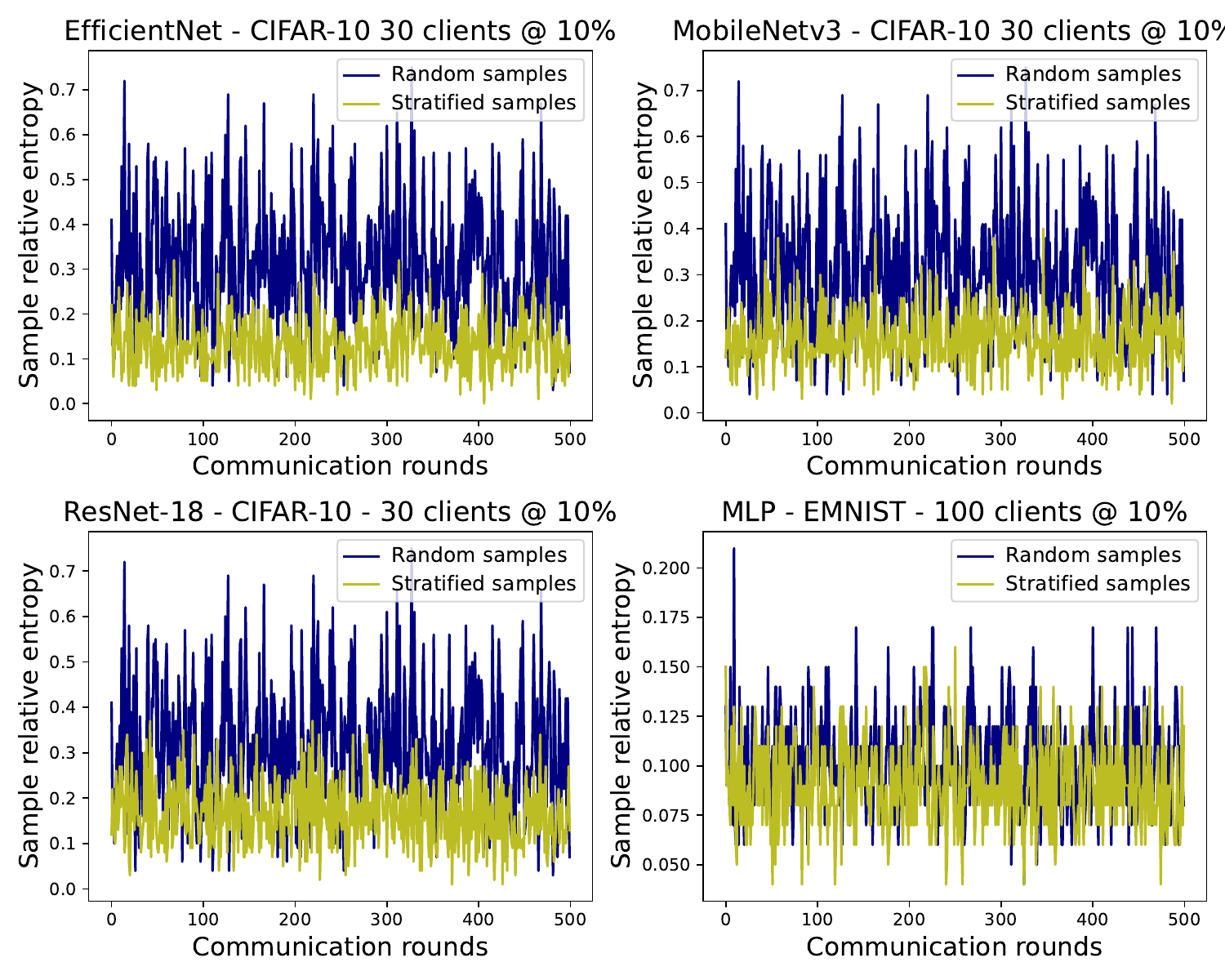}
    \caption{Comparing average relative entropy values of stratified samples vs. random samples over 500 communication rounds. The figure subtitles summarize the experiment settings (model - dataset - number of clients @ client sampling ratio)}
    \label{fig:entropies}
\end{figure*}

\begin{figure*}

  \includegraphics[width=1\linewidth]{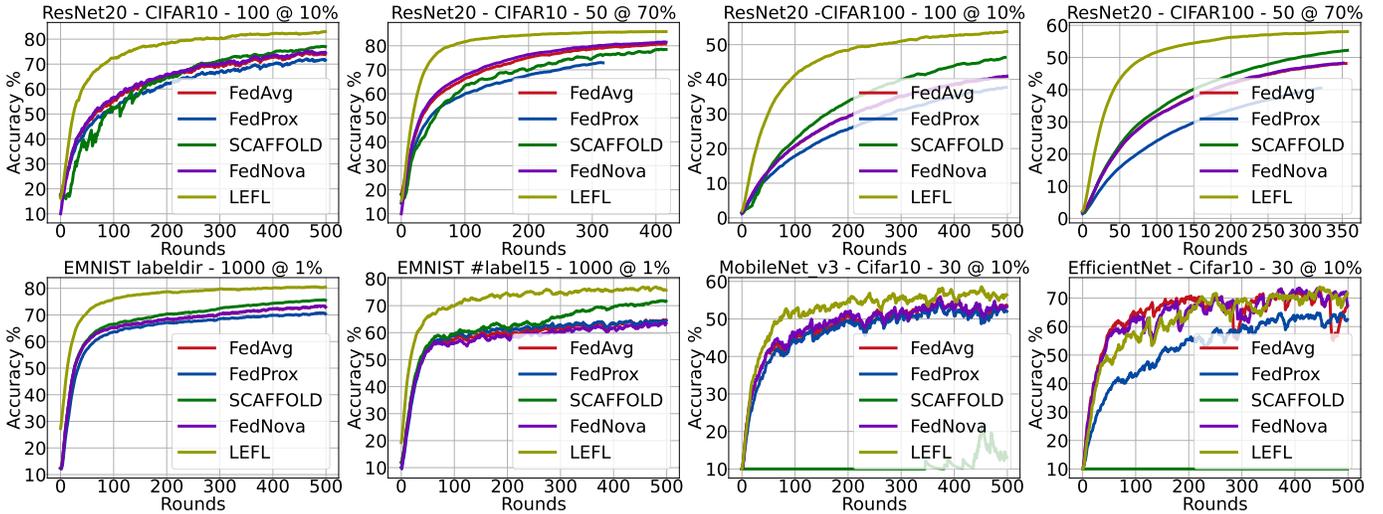}

\caption{Comparing performance of our work (LEFL) against FedAvg, FedProx FedNova and SCAFFOLD. Each experiment was conducted on a specified dataset using a predifined number of clients with a certain model and client sample ratio. The figure subtitles summarize the experiment settings (model - dataset - number of clients @ client sampling ratio).  }
\label{fig:compare}%
\end{figure*}

To test the conjectures, we designed a Federated Learning (FL) experiments involving $N$ participating clients. We partitioned the CIFAR-10 train data in a non-independent and identically distributed manner among the clients. In particular we implemented label-based skews where clients are assigned samples of labels from the Dirichlet distribution (distribution-based label skew) and where clients have strict number of labels selected at random (quantity-based label skew) as in \cite{li2022federated}. At the start of the experiment, clients  download a global model from the FL server and optimize it using their local dataset for 10 epochs. Afterwards, the clients use their model to make an inference on 1000 uniformly selected samples from a public dataset (TinyImageNet). The soft-label outputs from the client models is then used to construct a pairwise KL-based similarity matrix $M$, i.e., given models $m_i$ and $m_j$, $M[i][j] = \alpha^i_j = KL(p_i || p_j)$.

In line with conjecture 2, we set out to test whether similar rows $M[i], M[j]$ indicate models $m_i$ and $m_j$ have learned similar high level features. To test this hypothesis, we clustered the client models into $log(N)$ groups by feeding $M$ to a KMeans model. This step groups together clients with similar rows in the similarity matrix. Next we used 1000 samples of CIFAR-10 test data to compute the latent space vectors produced by the models as well as the Centered Kernel Alignment (CKA) \cite{kornblith2019similarity} values of the models for all layers. The latent space vectors were used to compute the average euclidean distance between the high level features of models within clusters. Similarly, CKA values were used to test similarities of learned features between inter-cluster and intra-cluster model pairs.

As shown in Fig.~\ref{fig:euclidean}, the average euclidean distance of latent space vectors of models clustered using this scheme is smaller compared to models clustered randomly. This indicates despite the stochastic distribution of data among client datasets, the proposed scheme performs better at clustering models that have learned similar high level features. Experiments A - D of Fig.~\ref{fig:euclidean} were conducted on the CIFAR-10 \cite{krizhevsky2009learning} dataset while E and F were conducted on the EMNIST \cite{cohen2017emnist} dataset. The details of the experiment will be included in the supplementary material.

Moreover, Fig.~\ref{fig:cka} juxtaposes CKA heatmaps for a pair of randomly selected inter-cluster models against a pair of intra-cluster models. The experiments involve 20 clients with CIFAR-10 dataset and ResNet-18 models grouped into 5 clusters. CKA intensity values range from 0 (no similarity) to 1 (complete similarity) values. Identical models are characterized by intense values across a heatmap's diagonal. The heatmaps suggest that while all pairs learned similar low level features (intense values in top left diagonal), the intra-cluster pairs additionally exhibit increased similarity in the latent layers of the model pairs (intense values in bottom right diagonal). In contrast, the inter-cluster pairs share a dark region in the latent layers. This indicates the utility of the similarity matrix in extracting similar models further reinforcing the proposed conjectures.

Finally, we measure the relative entropy of sampled client datasets by applying stratified sampling across the client clusters and selecting a proportional number of clients from each cluster. As shown in Fig.~\ref{fig:entropies}, the sampled clients dataset $D_s$ of the stratified samples maintains a relatively smaller relative entropy with respect to the global data distribution $D$ as compared to uniform client samples of the same size. This proves the proposed scheme offers a means of selecting clients with better representative local data in each FL round.

\subsection{Federated Learning Experiments}

We run several experiments on image classification datasets to test the performance of the client clustering scheme. In particular, we used the CIFAR-10, CIFAR-100, \cite{krizhevsky2009learning} and EMNIST datasets \cite{cohen2017emnist}.

CIFAR-10 is a popular computer vision dataset that consists of 60,000 32x32 color images in 10 classes, with 6,000 images per class. The dataset is divided into two sets, a training set of 50,000 images and a test set of 10,000 images. Similarly, the CIFAR-100 dataset is an advanced version of the CIFAR-10 dataset which also consists of 60,000 32x32 color images. It contains 100 classes with 600 images per class. The dataset is also split into a training set of 50,000 images and a test set of 10,000 images. Each class in the dataset contains exactly 500 training images and 100 testing images.

The EMNIST (short for Extended MNIST) dataset is a dataset of handwritten characters derived from the original MNIST dataset. It contains over 800,000 images, split into six different subsets. In our experiments, we consider the case where the dataset is split ``byclass'', which has a total of 814,255 images of handwritten characters from 62 classes, including uppercase and lowercase letters, digits, and special characters.

In all experiments, the datasets are partitioned in a non-iid manner. In particular, we utilize the benchmark from \cite{li2022federated} to apply distribution-based (labeldir) and quantity-based (labelX - where X is the number of unique labels per dataset) label skews. Distribution-based label skew divides the labels among clients, such that a proportion of the samples of each label is divided according to a Dirichlet distribution, i.e., each node $v$ is allotted  $p_{k,v} \sim Dir_N (\beta = 0.5)$ proportion of the instances of class $k$. Here, a concentration parameter $\beta \in [0,1]$ will determine the amount by which the labels are skewed. The smaller the value of $\beta$, the higher the non-iid skew.  The quantity-based label sets a hard limit on the number of labels each client possesses. In our EMNIST experiments for instance we specified skew value to label15 to specify that each client only possesses 15 labels out of a total of 62 labels found in the dataset.

In the CIFAR-10 experiments, each client is equipped with ResNet-20 \cite{he2016deep}, EfficientNet \cite{tan2019efficientnet} and MobileNet\_v3 \cite{howard2017mobilenets} models whereas in the EMNIST experiments clients deployed a 3-layer FFNN, with 784, 128 and 62 neurons. In each training round, the server samples a specified number of clients whose task is to download the latest global model, optimize it with their local datasets and upload the model parameters back to the server. For our proposed scheme, we utlize 1000 uniformly selected samples of TinyImagenet and FashionMNIST \cite{xiao2017fashion} as global data when conducting CIFAR and EMNIST experiments respectively. As such, the global data used to compute the similarity matrix is derived from a different distribution.  We examine the effect of our clustering scheme under different sampling ratios and evaluate both the convergence speed and accuracy as well as the communication cost to reach a specified target accuracy below.

\subsubsection{Convergence Analysis}

% Please add the following required packages to your document preamble:
% \usepackage{multirow}
% \usepackage{graphicx}
\begin{table*}[]
\centering
\caption{Comparing communication cost to reach a target accuracy of 70\% and 40\%.}
\resizebox{\textwidth}{!}{%
\begin{tabular}{cccccccc}

\hline
\multirow{2}{*}{\textbf{Dataset, Model}} & \multirow{2}{*}{\textbf{Target Accuracy}} & \multirow{2}{*}{\textbf{\begin{tabular}[c]{@{}c@{}}Algorithm\end{tabular}}}  & \multirow{2}{*}{\textbf{Communication Rounds}} & \multicolumn{3}{c}{\textbf{Communication Cost}}                                     \\
                                 &                                 &                                                                                     &                                   &                                                & \textbf{Round/Client} & \textbf{Total} & \textbf{$\Delta$ Cost}  \\ \hline
\multirow{5}{*}{\makecell{CIFAR-10 \\100 clients \\ 10\% sample ratio \\ResNet-20}}         &   \multirow{5}{*}{70\%}                      & FedAvg~\cite{mcmahan2017fedavg}                                                                             & 256    &                                                       & 2.1MB         & 5.2GB   & 0           \\
                                 &                        & FedProx~\cite{li2020federated}                                                                                     & 294                  &                                                            & 2.1MB                 & 6.02GB        & +0.82GB                   \\ 
                                 
                                 &                        & SCAFFOLD~\cite{karimireddy2020scaffold}                                                                                     &  234  &                                                                          & 4.3MB                 & 9.8GB        & +4.6GB                   \\
                                 
                                 &                        & FedNova~\cite{wang2020tackling}                                                                                     &   221      &                                                                   & 4.2MB                 & 9.06GB        & +3.86GB                   \\
                                 
                                 &                        & LEFL (Ours)                                                                                     &   \textbf{69}       &                                                                    & 2.1MB                 & 1.4GB + 208MB$^*$         & \textbf{-3.78GB}                   \\ \hline
\multirow{5}{*}{\makecell{CIFAR-100 \\100 clients \\ 10\% sample ratio \\ResNet-20}}         &   \multirow{5}{*}{40\%}                      & FedAvg~\cite{mcmahan2017fedavg}                                                                             & 421     &                                                        & 2.1MB         & 8.6GB   & 0           \\
                                 &                        & FedProx~\cite{li2020federated}                                                                                     &      >500       &                                                                  & 2.1MB                 & >10.2GB        & >+1.6GB                   \\ 
                                 &                        & SCAFFOLD~\cite{karimireddy2020scaffold}                                                                                     & 294   &                                                                         & 4.3MB                 & 12.3GB        & +3.7GB                   \\
                                 &                        & FedNova~\cite{wang2020tackling}                                                                                     &    419     &                                                                    & 4.2MB                 & 17.18GB        & +8.5GB                   \\
                                 &                        & LEFL (Ours)                                                                                     &  \textbf{79}   &                                                                          & 2.1MB                 & 1.62GB + 208MB$^*$        & \textbf{-6.9GB}                   \\ \hline
\multirow{5}{*}{\makecell{EMNIST \\1000 clients\\1\% sample ratio \\ 3-layer FFNN}}         & \multirow{5}{*} {70\%}                      & FedAvg~\cite{mcmahan2017fedavg}                                                                             & 224   &                                                          & 0.41MB         & 0.89GB   & 0             \\
                                 &                       &  FedProx~\cite{li2020federated}                                                                                    &   372     &                                                                     & 0.41MB                 & 1.48GB        & +0.59GB                   \\
                                 &                        & SCAFFOLD~\cite{karimireddy2020scaffold}                                                                                     &  174  &                                                                           & 0.83MB                 & 1.41GB        & +0.52GB                   \\
                                 &                        & FedNova~\cite{wang2020tackling}                                                                                     &   236   &                                                                       & 0.82MB                 & 1.88GB        & +0.99GB                   \\
                                 &                        & LEFL (Ours)                                                                                     &  \textbf{29}         &                                                                    & 0.41MB                 & 0.11GB + 0.8GB$^*$        & +0.02GB                   \\\hline

\multicolumn{7}{l}{$^*$ values show additional download/upload of public dataset and soft-labels required by our method.}

\end{tabular}%
}

\label{tab:com_cost}

\end{table*}

% \begin{table*}[]
% \centering
% \caption{Comparison of communication cost with SoTAs to achieve 80% accuracy.}
% \label{tab:com_cost}
% \resizebox{\textwidth}{!}{%
% \begin{tabular}{cccccccc}
% \hline
% \multirow{2}{}{\textbf{Method}} & \multirow{2}{}{\textbf{Model}} & \multirow{2}{}{\textbf{\begin{tabular}[c]{@{}c@{}}Target\ Accuracy\end{tabular}}} & \multirow{2}{}{\textbf{Clients}} & \multirow{2}{}{\textbf{Communication Rounds}} & \multicolumn{3}{c}{\textbf{Communication Cost (GB)}} & \multirow{2}{}{\textbf{$\Delta$ Cost}} \
% & & & & & \textbf{Round} & \textbf{Total} & \textbf{Per Client} & \
% \hline
% \multirow{2}{}{FedAvg~\cite{mcmahan2017fedavg}} & ResNet-20 & \multirow{3}{}{80%} & \multirow{2}{}{10} & 203 & 4.16 & 2.1 & 0 & 0 \
% & ResNet-32 & & & 192 & 6.00 & 3.2 & 0 & 0 \
% \hline
% \multirow{2}{}{FedNova~\cite{wang2020tackling}} & ResNet-20 & \multirow{3}{}{80%} & \multirow{3}{}{10} & 198 & 8.12 & 4.2 & 0.396 & +3.96 \
% & ResNet-32 & & & 197 & 12.31 & 6.4 & 0.631 & +6.31 \
% \hline
% \multirow{2}{}{FedProx~\cite{li2020federated}} & ResNet-20 & \multirow{3}{}{80%} & \multirow{3}{}{10} & 288 & 5.91 & 2.1 & 0.175 & +1.75 \
% & ResNet-32 & & & 400 & 12.80 & 3.2 & 0.680 & +6.80 \
% \hline
% \multirow{2}{}{(Ours)} & ResNet-20 & \multirow{3}{}{80%} & \multirow{3}{}{10} & 24 & XX & XX & XX & XX \
% & ResNet-32 & & & XX & XX & XX & XX & XX \
% \hline
% \end{tabular}%
% }
% \end{table*}

We run experiments for a specified number of rounds (eg. 500) to compare the convergence speed and convergence accuracies of the competing algorithms. All experiments used local training epochs $E_p = 10$.  Here, we wanted to see how the dataset complexity, data heterogeneity and sampling ratios affected convergence. As such, we experimented with different datasets with high (70\%), medium (10\%) and low (1\%) client sampling. Similarly, we experimented with both distribution-based and quantity-based label skews. We used $\beta = 0.5$ for our distribution-based skew and label15 for the quantity based skew, i.e, each client in this setting will possess a maximum of 15 labels.

In the ResNet-20 experiments, we used both CIFAR-10 and CIFAR-100 datasets. Each client used SGD with learning rate of 0.01 and weight decay rate of 0.99. We tested medium and high sample rates for clients of size 100 and 50 respectively. As shown in the first row of Fig.~ \ref{fig:compare}, our approach significantly improves the convergence speed while achieving an accuracy boost of 6\% in the CIFAR-10 100 clients experiment and 7.4\% in the CIFAR-100 100 client experiment. 

We conducted two experiments using the EMNIST dataset on 1000 clients and a sample rate of 1\% as shown in the first two plots of the second row in Fig.~ \ref{fig:compare}. We applied a distribution-based label skew (labeldir) in the first case and a quantity-based label skew (label15) in the latter. We used a 3-layer FFNN with an SGD optimizer, a learning rate of 0.01 and decay rate of 0.99. This experiment not only presents a small learning rate, but also a large number of participants which could potentially gain the most from stratified sampling. We observed an increase of 5\% and 6\% compared to SCAFFOLD for experiments one and two respectively.

Finally, we compared our scheme using MobileNet\_v3 \cite{MobilenetV3_Howard_2019_ICCV} and EfficientNet \cite{tan2019efficientnet} on the CIFAR-10 dataset. We used 30 clients and 10\% sample rate in all experiments. We kept the hyperparameters the same as the previous experiments. While the MobileNet\_v3 experiment yielded a 2.4\% increase, the EfficientNet experiments yielded a 1.3\% increase compared to FedNova. SCAFFOLD diverged in both cases as shown in the last two plots of the second row in Fig.~ \ref{fig:compare}. Overall, the EfficientNet and MobileNet\_v3 experiments showed a drop in performance for all algorithms due to increased model size which led to client models overfitting local data. In particular, while the ResNet-20 model we implemented was only about $\approx$1MB in size, the MobileNet\_v3 was $\approx$5MB and the EfficientNet was $\approx$15MB. Despite this, our method displayed marginal improvement.

\subsubsection{Communication Overhead}

In addition to improved convergence accuracy, our proposed scheme exhibits quicker convergence which translates into reduced communication overhead. To measure communication overhead, we count the number of rounds it takes the competing algorithms to achieve a specified target accuracy in Table \ref{tab:com_cost}. We also measure the communication cost ($\Delta$ Cost) relative to the FedAvg algorithm. This gives us a good idea of how much communication overhead all algorithms are incurring. 

We present three scenarios using the CIFAR-10, CIFAR-100 and EMNIST datasets. In the CIFAR experiments, we deploy 100 clients with ResNet-20 models and apply a client sample rate of 10\%. In the EMNIST experiment, we deploy 1000 clients using a 3-layer FFNN model and a client sample rate of 1\%. In all three cases, our proposed method achieves the target accuracy in significantly fewer number of rounds. Consequently, the proposed solution saved an extra 3.78GB and a 6.9GB worth of communication for the CIFAR-10 and CIFAR-100 experiments respectively. For the EMNIST experiments however, despite converging significantly faster, there was still a 0.02GB additional communication cost incurred as a result of the large number of clients and a relatively smaller model size. This cost was incurred when downloading the 1000 sample sized public dataset which was relatively large. Such scenarios will require a smaller public dataset, for instance, just having 100 samples. But in practice, this might not be an issue as FL models will likely be large deep learning models that are several times larger than the unlabeled public dataset.

\section{Ablation Study}

\begin{figure}

    \centering
    \includegraphics[width=1\linewidth]{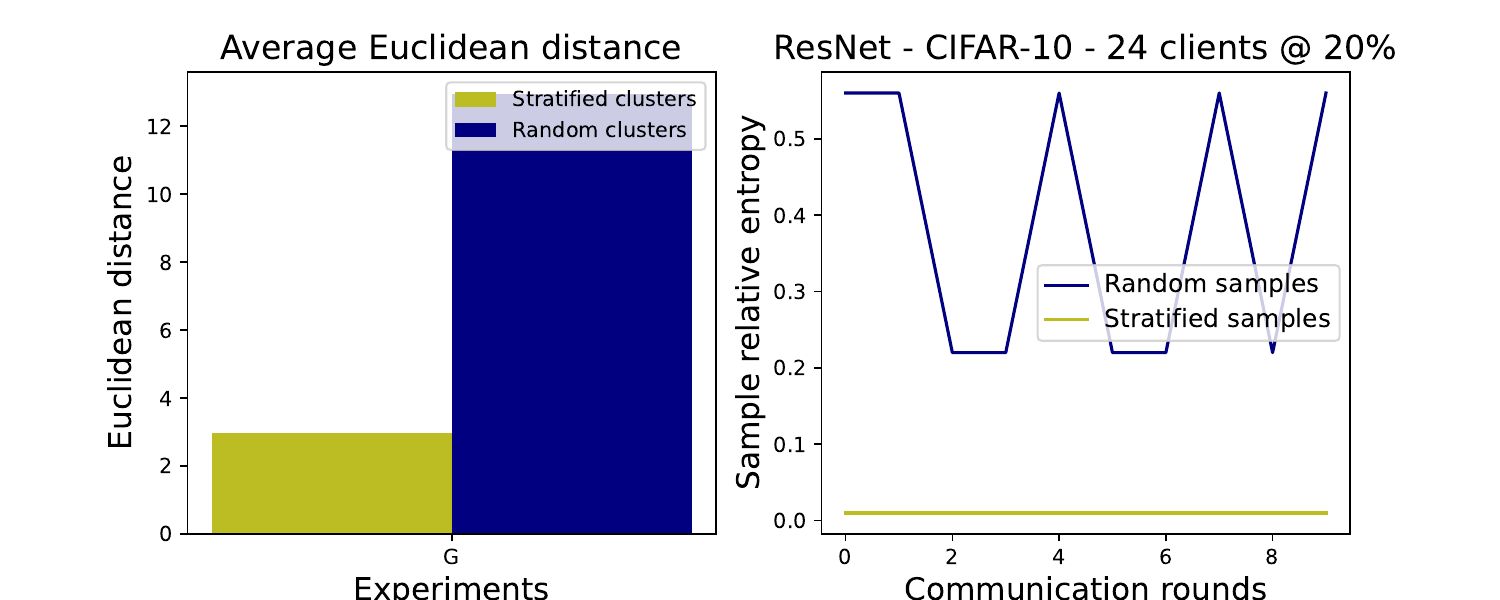}
    
    \caption{Comparing euclidean distance and average relative entropy for the manually partitioned quantity-based label skew. The relative entropy was measured for 10 rounds as opposed to the experiments in the main paper that were run for 500 rounds.}
    \label{fig:entropy_extra}
\end{figure}

We experimented with manual data split involving 24 models. Similar to the previous experiment, we applied a quantity based skew where the first two labels of the CIFAR-10 dataset were divided amongst 5 nodes, the next 2 labels among another 5 nodes, and so on, until the last pair of labels was divided among the remaining 4 nodes. As expected, we observed that the clustering scheme was able to group the clients that contained the same labels. The CKA values of two randomly selected model pairs in this experiment are presented in the last row of Fig. \ref{fig:cka} (i.e. Quantity-based label skew (manual)). The heatmaps exhibit a similar pattern as the examples in the main section. As shown in Fig. \ref{fig:entropy_extra} the average euclidean distance between the models inside the clusters vs models from random clusters has a wide gap. This can be attributed to the clustering scheme correctly classifying the models. The average relative entropy of our sampling strategy is also close to 0 whereas a random sampling strategy yields a much higher relative relative entropy.

\section{Limitations}

In this work, we make assumptions such as the uniformity of the participating client architectures and loss functions. Hence, these assumptions may not hold under system heterogeneity where client architectures vary. As such an alternative approach has to be taken such as utilizing knowledge distillation and a common knowledge network that serves as an intermediary among clients such as \cite{yu2022resource}.

\section{Conclusion}

As discussed in the introduction, the primary objective of this work is to implement a representative (stratified) client sampling in FL without compromising client data privacy. Selecting client samples with representative data reduces the chances of aggregating biased gradients and gradient drift. However, the server has no way of knowing client data, thus it cannot conduct stratified client sampling. In light of this, we proposed clustering clients based on their models learned high-level features. We empirically demonstrated that clustering models based on learned high-level features can help yield client samples with low relative entropy with respect to the global data distribution, i.e., it can improve the representativeness of the sampled clients dataset. However, computing the learned high-level features and comparing them against each other is an expensive task. A naive approach may require all clients to upload their models after the first FL round and make $n^2$ model comparisons. This, however, is impractical especially under a cross-device FL setting \cite{kairouz2021advances} which could possibly involve hundreds of thousands or even millions of client devices. 

In this paper, we first showed that clustering clients based on the learned high-level features can improve the representativeness of sampled client data. Next we proposed an affordable means of computing and comparing the learned high-level features via the soft-labels and similarity matrix. A small public dataset from a completely different distribution can be downloaded by all clients during initial contact. Similarly, uploading soft-labels back to the server is significantly cheaper than communicating the parameters of a large neural network. In Table \ref{tab:com_cost}, we show the extra upload/download cost incurred by the experiments are minimal.

Finally, the convergence experiments we conducted demonstrate that applying stratified client sampling using our method considerably improves the convergence and accuracy of the global model even against strong baselines like SCAFFOLD. Moreover, the number of rounds it takes to achieve a target accuracy is much fewer thus significantly decreasing the communication overhead and overall computational expense.

\bibliographystyle{plain}
\bibliography{bib}

\end{document}